# Continuous Sign Language Recognition Using Intra-Inter Gloss Attention

Hossein Ranjbar, Alireza Taheri[*]

Social and Cognitive Robotics Lab., Mechanical Engineering Department, Sharif University of Technology, Tehran, Iran

[*]Corresponding Author: artaheri@sharif.edu , Tel: +982166165531

**Abstract**

Many continuous sign language recognition (CSLR) studies adopt transformer-based architectures for sequence modeling due to their powerful capacity for capturing global contexts. Nevertheless, vanilla self-attention, which serves as the core module of the transformer, calculates a weighted average over all time steps; therefore, the local temporal semantics of sign videos may not be fully exploited. In this study, we introduce a novel module in sign language recognition studies, called intra-inter gloss attention module, to leverage the relationships among frames within glosses and the semantic and grammatical dependencies between glosses in the video. In the intra-gloss attention module, the video is divided into equally sized chunks and a self-attention mechanism is applied within each chunk. This localized self-attention significantly reduces complexity and eliminates noise introduced by considering non-relative frames. In the inter-gloss attention module, we first aggregate the chunk-level features within each gloss chunk by average pooling along the temporal dimension. Subsequently, multi-head self-attention is applied to all chunk-level features. Given the non-significance of the signer-environment interaction, we utilize segmentation to remove the background of the videos. This enables the proposed model to direct its focus toward the signer. Experimental results on the PHOENIX-2014 benchmark dataset demonstrate that our method can effectively extract sign language features in an end-to-end manner without any prior knowledge, improve the accuracy of CSLR, and achieve the word error rate (WER) of 20.4 on the test set which is a competitive result compare to the state-of-the-art which uses additional supervisions.

## 1. Introduction

Sign languages, as unique visual natural languages, serve as the primary means of communication for the hearing-impaired community. Sign languages are used by millions of people all over the world. In contrast to spoken languages, which rely on sound patterns, sign language conveys meaning through manual elements (e.g., hand configuration) and non-manual elements (e.g., facial expressions) [1] while it has its own distinct vocabulary and grammar. Therefore, mastering this language is rather difficult and time-consuming for hearing individuals, thus hindering direct communications between two groups.

Sign language recognition (SLR) acts as a bridge to address this gap. To alleviate this problem, isolated sign language recognition attempts to classify a video segment into an independent gloss[1]. Continuous sign language recognition (CSLR) takes this a step further by sequentially translating images into a series of glosses, enabling the expression of complete sentences and offering greater potential for real-world deployment.

Creating frame-level annotations is prohibitively expensive, leading most CSLR datasets to provide only sentence-level annotations. Consequently, researchers commonly approach video-based CSLR as a weakly supervised problem [2, 3]. Most CSLR models comprise three distinct components [4]: a spatial modeling module (2D-CNNs [4, 5], 3D-CNNs [6], or Vision Transformers [7]), a temporal modeling module (RNNs [6, 8], 1D-CNNs [2], or Transformer [5, 9, 10]), and an alignment module (CTC [5, 9, 10, 11] or hidden Markov models [12]). Figure 1 provides a simplified overview of the commonly used CSLR architectures.

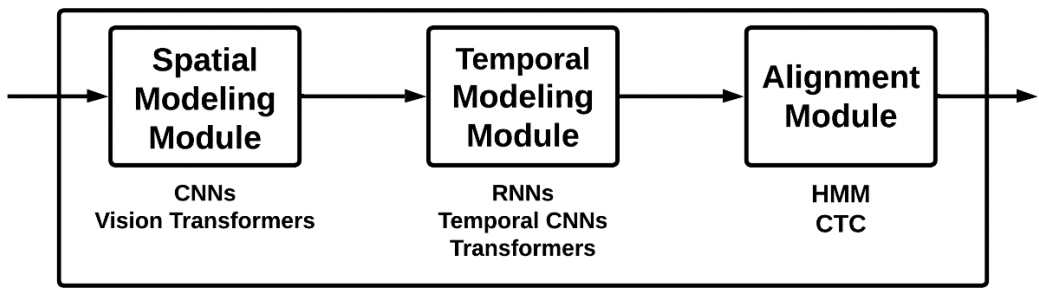

Figure 1. A simplified overview of commonly used CSLR architectures, consisting of three modules: the spatial modeling module, the temporal modeling module, and the alignment module.

The visual module's objective is to extract visual features from input videos. The realm of spatial modeling in computer vision was predominantly dominated by CNNs for an extended period. In recent times, the computer vision landscape has witnessed a significant transformation with the emergence of Vision Transformer (ViT) [13], marking a paradigm shift in spatial modeling. ViT is a pioneering effort to demonstrate how transformers can fully supplant standard convolutions in deep neural networks on large-scale image datasets. They applied the original Transformer model [14] with minimal modifications to a sequence of image 'patches' flattened as vectors. Since transformer architectures do not inherently encode inductive biases (prior knowledge) to deal with visual data, they typically demand a substantial amount of training data to figure out the underlying modality-specific rules. To address this limitation, several attempts have been made to include inductive biases (commonly used in CNNs) alongside the transformers to improve their efficiency [15, 16]. In CSLR, the most commonly used dataset, PHOENIX-2014 [17], includes less than 8,000 pairs of sign videos and gloss annotations. This quantity is insufficient to train a robust CSLR system with full supervision, as is typical in vision-language cross-modal tasks. Therefore, we employ a CNN-based model in the visual module, specifically utilizing the MobileNetV2 [18] architecture due to its low latency and high efficiency as a feature extractor.

---

[1] Gloss serves as the fundamental lexical unit for annotating sign languages.

Transformer-based architectures have been successfully adopted for numerous sequence modeling tasks, such as neural machine translation [14] and speech recognition [19], owing to their ability to effectively model global contexts. Thus, it is reasonable to introduce transformer to CSLR [5, 10] as well. However, in a sign language video, each gloss is typically short, spanning only a few frames (with an average of 12 frames per gloss in the PHOENIX-2014 dataset), underscoring the importance of local contexts. Vanilla self-attention, the core component of the transformer, computes a weighted average over all time steps, which may not fully exploit the local contexts and have a high computational cost.

In this paper, we introduce a novel framework that fully exploits local context while maintaining low computational cost. We present the intra-gloss attention module, which applies vanilla self-attention to chunks of the sign language video. Our objective is to extract relationships between frames within a window containing a gloss. Leveraging relationships between frames that are far apart is unnecessary, as they belong to different glosses. Additionally, we propose the inter-gloss attention module, which utilizes self-attention across different chunks (glosses) to capture the relationships between glosses that involve contextual and grammatical dependencies. Glosses are not chosen randomly; they have grammar and meaning relationships. For example, when 'cloud' (the translation of 'Wolke' in German) is used in a sentence, 'rain' (the translation of 'Regen' in German) is likely to be used in the same sentence. Both the intra- and inter-gloss attention modules (IIGA) are employed within the sequential module.

In the task of aligning input sequences with output sequences, Connectionist Temporal Classification (CTC) has achieved state-of-the-art performance on various tasks, including speech recognition [20], handwriting recognition [21], and computer vision tasks such as sentence-level lip reading [22], action recognition [23], hand shape recognition [24], and CSLR [5, 10]. We employ the CTC loss to train our model using weakly labeled gloss annotations. Additionally, we incorporate a preprocessing step in this architecture to help the model focus on the signer. We employ the MediaPipe [25] segmentation tool to remove the video background. The overall framework of the end-to-end CSLR proposed in this paper is shown in Figure 2.

The main contributions of this paper are as follows:

- We introduce a novel and efficient framework that includes both intra- and inter-gloss attention modules. The intra-gloss attention module captures frame relationships within chunks while reducing computational costs. The inter-gloss attention module applies self-attention across different chunks to capture contextual relationships between glosses.
- We have incorporated a segmentation step into this architecture to remove the background from the videos, enabling the model to focus on the signer.
- We conducted comprehensive experiments to illustrate the effectiveness of each component of our design. Despite our model having end-to-end training and not using additional factors like key point sequences and hand features, it has achieved competitive results in terms of Word Error Rate (WER) on the PHOENIX-2014 benchmark dataset.

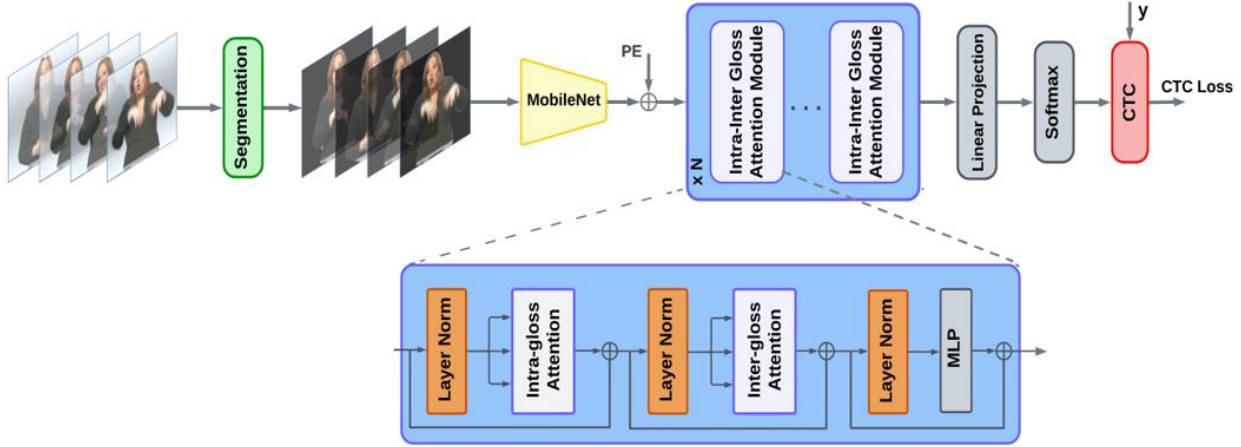

Figure 2. An overview of our proposed framework. The three main modules, namely spatial modeling (depicted in yellow), temporal modeling (in blue), and alignment (in red), are presented. In the initial step, a segmentation module is incorporated to remove the video background. It should be noted that ⊕ denotes the element-wise addition and N is the number of IIGA modules in the temporal modeling module.

## 2. Relates works

SLR tasks are generally categorized into three categories: finger-spelling recognition [26, 27], isolated word recognition [28, 29, 30], and continuous sign language recognition [2, 3, 4, 5]. In this paper, we concentrate on the latter category. CSLR aims to translate image frames into corresponding glosses in a weakly supervised manner, with only sentence-level labels provided. Earlier methods in CSLR always employed hand-crafted features or HMM-based systems [17, 31] to perform temporal modeling and translate sentences step by step. In recent years, with the rapid development of deep learning, the network structures of "CNN+RNN" and "CNN+Transformer" have become the prevailing choices for CSLR. For example, SLT [10] introduced a transformer-based architecture that simultaneously learns continuous sign language recognition and translation while being trainable in an end-to-end manner. This marked one of the early instances of a transformer-based architecture in the field of CSLR.

Despite the success of vision transformer models in high-level vision problems [13, 32, 33], they have not been extensively adopted in CSLR visual modules. Ref. [7] is one of the few efforts that use ViT. It mostly focuses on extracting features and introduces a feature extraction network that can simultaneously learn the spatial-temporal representation of RGB and skeleton data. It used the ViT to extract RGB clips' spatial-temporal features and designed an attention-enhanced multi-scale 3D graph convolutional network (AM3D-GCN) as the feature extractor for the skeleton clips.

Given the importance of exploiting the local context to enhance the model's performance, some efforts leverage local context in either the spatial or temporal domain [5, 9]. To exploit spatial local context, the authors of [9] introduced SAN (Sign Attention Module) as an end-to-end framework for CSLR that utilizes vanilla self-attention for temporal modeling. It also integrates an additional branch to incorporate handshapes with their spatiotemporal context. To exploit temporal local context, the authors of [5] propose a self-attention mechanism with local contextual information at two levels of model computation to leverage both global and local context for sequential modeling. At the score level, the attention scores are modulated directly with an additional Gaussian bias to weaken the relations between distant query-key pairs; and at the query level, (since the projection of queries in vanilla self-attention is performed position-wise,) a depth-wise temporal convolutional network layer is attached to the projection to get local context-aware queries. Contrary to the capability of this method to leverage both global and local contexts, it computes a similarity score between all pairs of input features, which has a significant computational cost. These inspire us to introduce the IIGA module in sequential modeling to leverage local context and extract dependencies between the glosses. To the best of our knowledge, no previous study has considered doing this for SLR.

## 3. Methodology

### 3.1. Framework Overview

As shown in Figure 1, the backbone of CSLR models is composed of a visual module, a sequential module, and an alignment module. Additionally, our architecture includes a segmentation module designed to remove the background from the images. Figure 2 provides an overview of our proposed framework. Given a video with T frames $X = \{x_i\}_{i=1}^{T} \in \mathbb{R}^{T \times C \times H \times W}$, where C = 3 denotes the dimension of the RGB images and H and W represent the frame height and width, respectively, our visual module initially extracts the visual features $Z = \{z_i\}_{i=1}^{T} \in \mathbb{R}^{T \times D}$, where D is the dimension of the embedded vector. This is performed frame by frame using MobileNet-V2 [18]. Subsequently, the sequential module further extracts the sequential features $S = \{s_i\}_{i=1}^{T} \in \mathbb{R}^{T \times D}$. In this module, we utilize the IIGA module. Finally, in the alignment module, we utilize CTC [23] to compute the probability of the gloss label sequence $p(x|y)$, where $Y = \{y_i\}_{i=1}^{N}$ and N is the length of the label sequence.

### 3.2. Segmentation

Sign languages convey information mainly through signers' facial expressions and hand gestures [34]. Thus, we hope the visual module can focus on these informative regions. Furthermore, the interaction between the signer and the background is of minor significance. Motivated by this idea, we integrate a segmentation module that removes the background of the images. We employ Mediapipe-Holistic, which includes a segmentation tool. This tool generates a probability mask indicating the presence of a person. Where the probability of a person's presence is higher, the value of the corresponding element of the mask gets closer to 1; and where the probability of a person's presence is lower, the value of the corresponding element of the mask gets closer to zero. By applying this mask to the target image through multiplication, we effectively remove the background:

$$X' = F_{seg}(X) \in \mathbb{R}^{T \times C \times H \times W} \tag{1}$$

$$x'_i = x_i \times M_i \tag{2}$$

Where $F_{seg}$ is the segmentation module, and $M_i$ is the mask of each frame. Adding this module could help the visual module focus on informative region areas, particularly in the case of datasets featuring complex backgrounds.

### 3.3. Visual Module

In this module, CNNs (2D CNNs and 3D CNNs) have traditionally been a common choice. However, more recently, transformer-based architectures have gained popularity as well. We opt for CNNs over Vision Transformer architectures in this context, primarily due to the fact that vision transformers lack inherent encoding of inductive biases and have shown comparatively poorer performance on small-scale datasets when compared to CNNs. We utilized MobileNet-V2 for several compelling reasons. First and foremost, MobileNet-V2's inherent efficiency and lightweight design render it exceptionally suitable for deployment on resource-constrained devices. Furthermore, MobileNet-V2 exhibits a favorable trade-off between model complexity and performance, offering us the capability to achieve commendable accuracy while respecting computational limitations. The architecture's inherent scalability provides us with the flexibility to adjust model complexity to the specific demands of our task. By capitalizing on transfer learning with pre-trained MobileNet-V2 models, we can leverage the knowledge gained from extensive datasets and fine-tune the network to achieve outstanding results, even with limited available data. Lastly, MobileNet-V2's low inference latency ensures the timely and responsive execution of our model, which is paramount for our application. After removing the background from the frames, an ImageNet pre-trained model of MobileNet-V2 is applied to them to extract spatial features:

$$Z = f_{conv}(X') \in \mathbb{R}^{T \times D} \tag{3}$$

Where $f_{conv}$ is the MobileNet-V2 and D represents a feature dimension and is equal to 1280.

### 3.4. Sequential Module

The transformer network further takes $Z$ as input. At the heart of the transformer architecture lies self-attention [14], a concept we briefly introduce here to ensure the paper's self-contained nature. In essence, self-attention computes a weighted average of features, with the weight proportional to a similarity score between pairs of input features.

Given the feature sequence $Z = \{z_i\}_{i=1}^{T} \in \mathbb{R}^{T \times D}$ with T time steps of D dimensional features, Z is projected using $W_Q \in \mathbb{R}^{D \times D_q}$, $W_K \in \mathbb{R}^{D \times D_k}$, and $W_V \in \mathbb{R}^{D \times D_v}$ to extract feature representations $Q$, $K$, and $V$, referred to as query, key and value, respectively with $D_k = D_q$. The outputs $Q, K, V$ are computed as

$$Q = Z W_q \quad K = Z W_k \quad V = Z W_v \tag{4}$$

The output of the self-attention mechanism is determined by the following computation:

$$S = softmax\left(\frac{QK^T}{\sqrt{D_q}}\right)V \tag{5}$$

Here, $S \in \mathbb{R}^{T \times D}$, and the row-wise softmax operation is performed. Multi-headed self-attention (MSA) extends the concept by incorporating multiple self-attention operations in parallel. An inherent advantage of MSA lies in its capacity to incorporate temporal context across the entire sequence, yet such a benefit comes at the cost of computation. A vanilla MSA has a complexity of $O(T^2D + D^2T)$ in both memory and time, making it highly inefficient for processing long videos [35]. In the vanilla MSA, relations between query-key pairs are calculated among all embedded vectors, incurring a high computational cost. The primary objective of employing the self-attention mechanism in CSLR is to establish connections between gloss frames. Given that there is no significant relationship between frames that are distant from each other, obtaining query-key pair relations over long distances seems to be ineffective. In addition, considering irrelevant frames in the self-attention mechanism may introduce extra noise to the gloss segment modeling process.

### 3.4.1. Intra-gloss attention

There have been several recent studies on local context in self-attention [35, 36]. Here we propose the intra-gloss attention module that performs self-attention within each chunk. Employing the inter-gloss attention significantly reduces the complexity to $O(W^2TD + D^2T)$, where W represents the chunk size. In this module, the embedded vector of the video is segmented into overlapping chunks of uniform length; and the attention mechanism is then applied within each of these chunks. As illustrated in Figure 3, each color denotes a chunk, and self-attention is applied to each of them. An attention score $M \in \mathbb{R}^{w \times w}$ represents dependencies between the chunk frames.

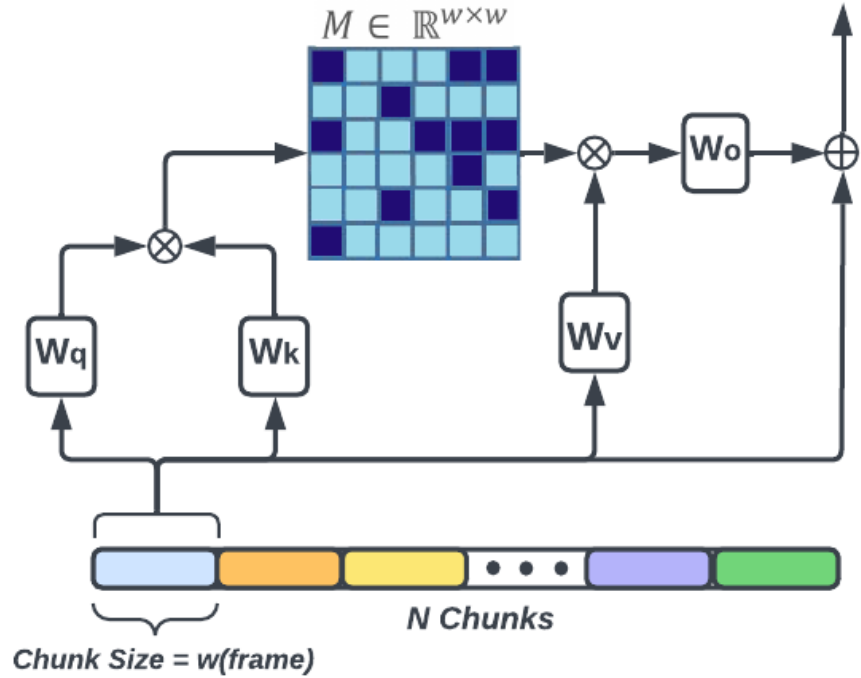

Figure 3. The intra-gloss attention module. Self-attention is applied within each chunk. For the sake of better visual representation, the chunks are intentionally not overlapping.

**3.4.2. Inter-gloss attention**

The glosses used in the video were not chosen by chance and have a semantic connection. Undoubtedly, identifying this semantic connection and dependence between glosses can increase the performance of the model. For instance, when 'cloud' ('Wolke' in German) is used in a sentence, it is likely that 'rain' ('Regen' in German) will also be used in the same sentence. This module follows the intra-gloss attention module, and the outputs of the intra-gloss attention module are processed through normalization. We utilize the self-attention mechanism to capture relationships across multiple gloss chunks. We begin by aggregating the features at the chunk level within each chunk through temporal dimension averaging. Subsequently, we apply multi-head self-attention to all chunk-level features to capture interactions between different chunk pairs. The resulting features are replicated along the time axis and added to the original feature in a residual manner. Figure 4 illustrates the inter-gloss attention module.

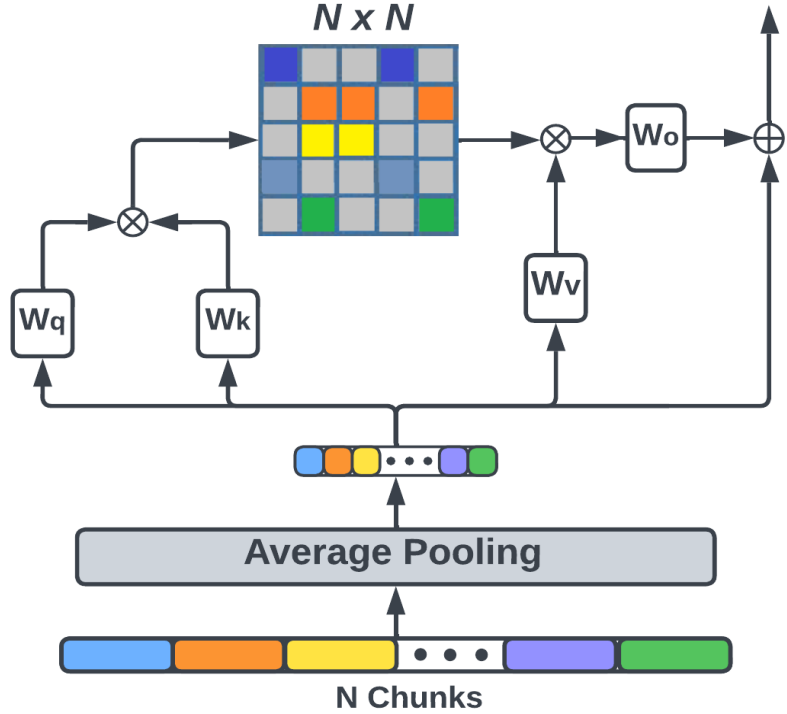

Figure 4. The Inter-gloss attention module. The self-attention mechanism is applied to all chunk-level features.

### 3.5. Alignment Module and Loss Function

To facilitate end-to-end training using weakly annotated data, Ref. [23] introduced the CTC loss function. This loss function accounts for all potential alignments between two sequences when computing the error, and it rapidly gained popularity in a wide range of sequence-to-sequence applications. It produces a label for each time step, which can be either a repeating label or a special blank symbol. Under the assumption of conditional independence, for a given input sequence x, the conditional probability of a label sequence $\emptyset = \{\emptyset_i\}_{i=1}^{N}$, where $y_i \in G \ \{blank\}$ and $G$ is the gloss vocabulary, can be estimated by:

$$p(\emptyset|X) = \prod_{i=1}^{T} p(\emptyset_i|X) \tag{6}$$

where $p(y_i|X)$ represents the frame-level gloss probabilities generated by a classifier. Ultimately, the probability of yielding the true label sequence is the sum of all possible alignments:

$$p(y|x) = \sum_{\emptyset = \varphi^{-1}(y)} p(\emptyset|x) \tag{7}$$

Here, $\varphi$ represents a mapping function employed to eliminate duplicate and blank symbols $\emptyset$. The CTC loss can then be defined as:

$$L_{CTC} = -\log p(x|y) \tag{8}$$

## 4. Experiments

### 4.1. Dataset and Evaluation

#### 4.1.1. Dataset

PHOENIX-2014 is a publicly available German Sign Language dataset, which is the most popular benchmark for CSLR. The corpus was recorded from broadcast news about the weather. The dataset contains a total of 963k frames captured by an RGB camera at a rate of 25 frames per second. It has a vocabulary size of 1295. There are 5672, 540, and 629 data samples in the train, evaluation, and test sets, respectively. The dataset contains 9 signers who appear in all three splits.

#### 4.1.2. Evaluation Metrics

In CSLR, word error rate (WER) stands as the most commonly employed metric for performance evaluation. It measures the number of necessary insertions, substitutions, and deletions in the recognized sentences when compared to the reference sentences. WER is an edit distance, indicating the least number of operations of substitutions (#sub), insertions (#ins), and deletions (#del) to transform the predicted sentences into the reference sequences (#reference):

$$WER = \frac{\#sub + \#del + \#ins}{\#reference} \tag{9}$$

### 4.2. Implementation Details

#### 4.2.1. Data Augmentation

For training, we implement stochastic frame dropping [4], using a dropping ratio of 0.5. During this process, we randomly select 50% of the frames from the original video clip. During testing, an equal 50% of the frames are uniformly chosen from the video. This approach effectively reduces computational complexity and eliminates frame redundancy. The videos are resized to $256 \times 256$ and subsequently cropped to $224 \times 224$. During training, we apply random cropping and random affine (10 degrees) for data augmentation. During testing, we solely utilize center cropping.

#### 4.2.2. Backbone and Hyper-parameters

The images are normalized by subtracting the dataset's image mean. For spatial image embeddings, we utilize the MobileNetV2 architecture [18], where we omit the last fully connected layers and employ the remaining layers for feature embeddings. Each input frame is encoded into a 1280-dimensional vector, and the hidden size of the IIGA module is set to 1280.

In the sequential module, we incorporate two IIGA modules with 8 heads (h = 8), the model dimension (hidden size) of d = 1280, and the position-wise feed-forward layer dimension of $d_{ff}$ = 2048. Also, relative positional encoding is used in this module.

For the intra-gloss attention module, the chunk size is set to 12 (this configuration yields the best results). Additionally, following the IIGA module, we introduce a layer for the position-wise feed-forward network.

### 4.2.3. Training

All components of our network were constructed using PyTorch [37]. For initialization, we employ Xavier initialization [38] for all network layers, except for the image embedding layers, which have been pre-trained on the ImageNet [39]. We utilize the Adam [40] optimization method with its default parameters, a learning rate of 1.0e−4, and a weight decay factor of 1.0e−5. Additionally, we implement a multi-step learning rate schedule, reducing the learning rate by a factor of 0.1 at epoch 15. The dropout rate is set to 0.1 to prevent overfitting. We also use gradient clipping with a threshold of 1. In our training process, we use a small batch size of 2. Since we are performing batch training and considering that input clips may vary in size, sequences within each batch are padded to equal lengths, which corresponds to the maximum sequence length within that batch. This necessitates the use of a mask on the input sequences to prevent attention on the padded elements.

We train our networks for 30 epochs or until train perplexity convergence. Model evaluation is performed every epoch on the Dev (validation) set, and we report the best-performing model. Our training process is conducted on a single NVIDIA Quadro RTX 8000 GPU with 48GB of memory.

### 4.3 Ablation Study

### 4.3.1. Effectiveness of Segmentation

We anticipated that by incorporating the segmentation, the model would focus on the signer, resulting in improved performance. Segmentation introduces additional computational costs and diminishes inference speed, yet it notably reduces the word error rate. The experimental results are shown in Table 1. Considering that PHOENIX-2014 dataset videos have a relatively simple background, the impact of this module becomes more pronounced when applied to datasets with more complex backgrounds.

Table 1. Study for the effectiveness of segmentation

| Segmentation | WER (%) | |
| --- | --- | --- |
| | Dev | Test |
| ✗ | 21.5 | 21.9 |
| ✓ | 20.1 | 20.4 |

**4.3.2. Effectiveness of Data Augmentation**

Given the limited amount of training data in CSLR, data augmentation becomes crucial. In this study, we employ data augmentations, as detailed in Section 4.2, which encompasses stochastic frame dropping, random cropping, and random affine transformations. In this study, we investigate the effectiveness of augmentation and demonstrate its capacity to prevent overfitting and enhance accuracy, as shown in Table 2.

Table 2. Ablation for effectiveness of augmentation. The best result has been shown in bold.

| Augmentations | | | WER (%) | |
| --- | --- | --- | --- | --- |
| Stochastic frame dropping | Random Cropping | Random Affine | Dev | Test |
| - | - | - | 21.2 | 21.5 |
| ✓ | | | 20.8 | 20.9 |
| ✓ | ✓ | | 20.3 | 20.5 |
| ✓ | ✓ | ✓ | **20.1** | **20.4** |

**4.3.3 Effectiveness of Intra-Inter Gloss Attention**

As shown in Table 3, it is clear to see that both intra-gloss and inter-gloss attention modules can improve the performance of the model. As previously mentioned, the implementation of local attention serves to mitigate complexity and reduce computational costs. Moreover, this approach is anticipated to lead to a decrease in WER as the attention mechanism operates within chunks. As indicated in Table 3, the error rate has decreased by 0.5% when compared to the base model. Furthermore, we observed a 20% increase in the inference speed.

We study the effects of the chunk size of the intra-gloss attention module in our model. We vary the chunk size, retrain the model, and present the model WER in Table 4. Reducing the chunk size cuts down the FLOPs while also decreasing the WER. Excessive reduction in chunk size can lead to the loss of valuable frame information. The best performance was achieved with a chunk size of 12. In the PHOENIX-2014 dataset, the average number of frames per word is 12, which likely explains why a chunk size of 12 yields the optimal results.

Table 3. Ablation for the effectiveness of intra-gloss and inter-gloss attention modules. The best result has been shown in bold.

| Models | WER (%) | | #Param. (M) |
|---|---|---|---|
| | Dev | Test | |
| Vanilla self-attention | 21.3 | 21.7 | 29.68 |
| Intra-gloss attention | 20.7 | 21.1 | 29.68 |
| **Intra-inter gloss attention** | **20.1** | **20.4** | **42.81** |

Table 4. The impact of chunk size on Model Accuracy. The best result has been shown in bold.

| Chunk Size | WER (%) | |
|---|---|---|
| | Dev | Test |
| 10 | 20.5 | 21.0 |
| **12** | **20.1** | **20.4** |
| 14 | 20.4 | 20.8 |
| 16 | 21.1 | 21.4 |

Incorporating the inter-gloss attention module, as opposed to the intra-gloss attention module, raises the complexity and augments the number of network parameters, potentially leading to overfitting. The reduction in the WER is more pronounced during training compared to testing.

**4.3.4. Hyper-parameters**

Ablation studies were conducted to fine-tune the hyper-parameters of our proposed model. As shown in Table 5, the best result is obtained with the following hyper-parameters: MobileNet-V2 embedding network, a batch size of 2, two encoder layers, and a dropout rate of 0.1. In one experiment, we utilized the ResNet18 [41] network to extract spatial features; but we found that the MobileNet-V2 outperformed it in terms of performance. When the number of encoder layers is set to three, the number of the model's parameters (count) rises to 41 million, increasing its complexity. This, in turn, can lead to overfitting, resulting in a higher word error rate. The optimal dropout is 0.1, while a dropout value of 0.3 makes the training process more challenging and leads to a slower convergence.

Table 5. Ablation study for tuning the hyper-parameters. The best result has been shown in bold.

| CNN | Batch Size | Encoder Layers | Dropout | WER (%) Dev | Test |
|---|---|---|---|---|---|
| ResNet18 | 1 | 2 | 0.3 | 26.2 | 26.7 |
| MobileNet-v2 | 1 | 2 | 0.3 | 25.7 | 26.3 |
| MobileNet-v2 | 2 | 2 | 0.3 | 23.2 | 23.6 |
| MobileNet-v2 | 2 | 3 | 0.3 | 24.8 | 25.3 |
| **MobileNet-v2** | **2** | **2** | **0.1** | **20.1** | **20.4** |

In Figure 5, our model's predictions of two example videos of PHOENIX-2014 test set are shown.

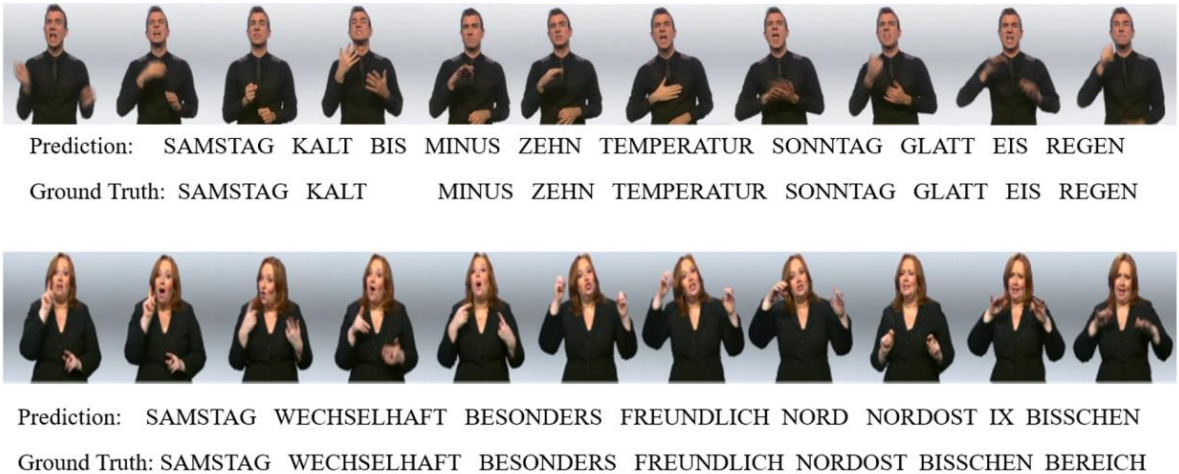

Figure 5. Our model's predictions and the ground truth of two sign videos in the test set of the PHOENIX-2014 [17].

### 4.3.5. Comparison with the State-of-the-art Results

Table 6 provides a comprehensive comparison between our approach and other methods on the PHOENIX-2014 dataset. Notably, our proposed model is fully end-to-end trainable and does not depend on any prior knowledge, such as skeleton data or hand and face cropping patches. Despite these advantages, our model achieves competitive results and outperforms models that exclusively utilize the RGB frames. The current state-of-the-art on the PHOENIX-2014 test set is TwoStream-SLR [44] with a WER of 18.8. This model incorporates additional supervision by integrating pose data and treating the averaged ensemble predictions as pseudo-targets. Furthermore, it introduces an additional self-distillation loss, which is computed per frame to facilitate training with extra pseudo frame-level supervision. Notably, our model, without any auxiliary loss or additional factors such as pose, hand, and face features, achieves competitive results.

Table 6. Comparison with methods on PHOENIX-2014.

| Method | More supervision | | | | | End-to-end | WER | |
|---|---|---|---|---|---|---|---|---|
| | Full | Pose | Hand | Face | Flow | | Dev(%) | Test(%) |
| SubUNets [24] | ✓ | | ✓ | | | ✓ | 40.8 | 40.7 |
| DenseTCN [42] | ✓ | | | | | ✓ | 35.9 | 36.5 |
| CNN-LSTM-HMMs [12] | ✓ | | ✓ | | | ✗ | 26.0 | 26.0 |
| DNF (RGB) [3] | ✓ | | | | | ✗ | 23.8 | 24.4 |
| DNF (RGB+Flow)[3] | ✓ | | | | ✓ | ✗ | 23.1 | 22.9 |
| VAC [42] | ✓ | | | | | ✓ | 21.2 | 22.3 |
| LCSA [5] | ✓ | | | | | ✓ | 21.4 | 21.9 |
| STMC-R [34] | ✓ | ✓ | ✓ | ✓ | | ✓ | 21.1 | 20.7 |
| C²SLR [43] | ✓ | ✓ | | | | ✓ | 20.5 | 20.4 |
| TwoStream-SLR [44] | ✓ | ✓ | | | | ✓ | **18.4** | **18.8** |
| **IIGA (Ours)** | ✓ | | | | | ✓ | **20.1** | **20.4** |

## 5. Limitations and future work

CSLR faces challenges due to insufficient data; the benchmark dataset, PHOENIX-2014, comprises fewer than 8,000 videos. This scarcity makes training a deep network for this intricate, multi-channel task challenging. For future work, we plan to apply the intra-inter gloss attention module to Iranian sign language translation. Additionally, we aim to integrate this module with other transformer-based modules capable of fully leveraging both local and global contexts. Exploring the use of pose data instead of RGB images is a challenging trade-off, significantly reducing computational costs but potentially losing some information. We intend to employ the proposed model in a new configuration for skeleton-based CSLR.

## 6. Conclusion

In this paper, we propose a novel IIGA module for CSLR. In this work, to address the importance of local context in sequential modeling and reduce model complexity and computational costs, we introduce the intra-gloss attention module, which in addition to increasing the inference speed by 20%, also increases the accuracy of the model. To address the importance of dependencies between glosses in the sentence, we introduce the inter-gloss attention module. This module helps the model capture these dependencies and interactions among glosses, thereby improving recognition. In SLR, the hand and face regions are more informative, while the interaction between the signer and the environment is not as critical. To address this, we add segmentation to the model to remove the background from the videos. Extensive ablation studies validate the effectiveness of intra-inter gloss attention and segmentation. More remarkably, this model achieves competitive results in terms of WER on the PHOENIX-2014 benchmark dataset since it is end-to-end and solely based on the RGB frames.

**Acknowledgment**

This study was funded by the "Sharif University of Technology, Iran" (Grant No. 980517). We would also like to appreciate "Dr. Ali Akbar Siassi Memorial Research Grant Award" for the support of our projects at the Social and Cognitive Robotics Lab. For the English editing of the final manuscript, we have used Chat-GPT.

**Conflict of interest**

Author Alireza Taheri received a research grant from the "Sharif University of Technology" (Grant No. G980517). The author Hossein Ranjbar declares that he has no conflict of interest.

**Availability of data and material (data transparency)**

The main used dataset is available for researchers worldwide. All data from this project is available in the Social & Cognitive Robotics Laboratory archive.

**Code availability:**

All of the codes are available in the Social & Cognitive Robotics Laboratory archive. If the readers need the codes, they may contact the corresponding author.

**Authors' contributions:**

Both authors contributed to the study's conception and design. Material preparation and analysis were performed by Hossein Ranjbar. The first draft of the manuscript was written by Hossein Ranjbar, and both authors commented on the manuscript. Both authors read and approved the final manuscript. Alireza Taheri supervised the study.

**Ethical Approval**

There were no tests on humans/animals in this study. We used one of the worldwide available datasets to conduct this study.

**Consent to participate:**

In this study, we did not directly involve any participants and used one of the worldwide available datasets. It seems that the informed consent was previously obtained from all individual participants.

**Consent for publication:**

We have not used any participants' images in this manuscript.

**References**

[1] Ong SC, Ranganath S. Automatic sign language analysis: a survey and the future beyond lexical meaning. IEEE Trans Pattern Anal Mach Intell. 2005 Jun;27(6):873-91. doi: 10.1109/TPAMI.2005.112. PMID: 15943420.

[2] Cheng, K. L., Yang, Z., Chen, Q., & Tai, Y. W. (2020). Fully convolutional networks for continuous sign language recognition. In *Computer Vision–ECCV 2020: 16th European Conference, Glasgow, UK, August 23–28, 2020, Proceedings, Part XXIV 16* (pp. 697-714). Springer International Publishing.

[3] Cui, Runpeng, Hu Liu, and Changshui Zhang. "A deep neural framework for continuous sign language recognition by iterative training." *IEEE Transactions on Multimedia* 21.7 (2019): 1880-1891.

[4] Niu, Z., & Mak, B. (2020). Stochastic fine-grained labeling of multi-state sign glosses for continuous sign language recognition. In *Computer Vision–ECCV 2020: 16th European Conference, Glasgow, UK, August 23–28, 2020, Proceedings, Part XVI 16* (pp. 172-186). Springer International Publishing.

[5] Zuo, R., & Mak, B. (2022). Local Context-aware Self-attention for Continuous Sign Language Recognition}}. *Proc. Interspeech 2022*, 4810-4814.

[6] Pu, J., Zhou, W., & Li, H. (2019). Iterative alignment network for continuous sign language recognition. In *Proceedings of the IEEE/CVF conference on computer vision and pattern recognition* (pp. 4165-4174).

[7] Li, R., & Meng, L. (2022). Multi-view spatial-temporal network for continuous sign language recognition. *arXiv preprint arXiv:2204.08747*.

[8] Pu, J., Zhou, W., Hu, H., & Li, H. (2020, October). Boosting continuous sign language recognition via cross modality augmentation. In *Proceedings of the 28th ACM International Conference on Multimedia* (pp. 1497-1505).

[9] Slimane, F. B., & Bouguessa, M. (2021, January). Context matters: Self-attention for sign language recognition. In *2020 25th International Conference on Pattern Recognition (ICPR)* (pp. 7884-7891). IEEE.

[10] Camgoz, Necati Cihan, et al. "Sign language transformers: Joint end-to-end sign language recognition and translation." *Proceedings of the IEEE/CVF conference on computer vision and pattern recognition*. 2020.

[11] Zheng, J., Wang, Y., Tan, C., Li, S., Wang, G., Xia, J., ... & Li, S. Z. (2023). Cvt-slr: Contrastive visual-textual transformation for sign language recognition with variational alignment. In *Proceedings of the IEEE/CVF Conference on Computer Vision and Pattern Recognition* (pp. 23141-23150).

[12] Koller, O., Camgoz, N. C., Ney, H., & Bowden, R. (2019). Weakly supervised learning with multi-stream CNN-LSTM-HMMs to discover sequential parallelism in sign language videos. *IEEE transactions on pattern analysis and machine intelligence*, *42*(9), 2306-2320.

[13] Dosovitskiy, A., Beyer, L., Kolesnikov, A., Weissenborn, D., Zhai, X., Unterthiner, T., ... & Houlsby, N. (2010). An image is worth 16x16 words: Transformers for image recognition at scale. arXiv 2020. *arXiv preprint arXiv:2010.11929*.

[14] Vaswani, A., Shazeer, N., Parmar, N., Uszkoreit, J., Jones, L., Gomez, A. N., ... & Polosukhin, I. (2017). Attention is all you need. *Advances in neural information processing systems*, *30*.

[15] Esser, P., Rombach, R., & Ommer, B. (2021). Taming transformers for high-resolution image synthesis. In *Proceedings of the IEEE/CVF conference on computer vision and pattern recognition* (pp. 12873-12883).

[16] Li, Y., Zhang, K., Cao, J., Timofte, R., & Van Gool, L. (2021). Localvit: Bringing locality to vision transformers. *arXiv preprint arXiv:2104.05707*.

[17] Koller, O., Forster, J., & Ney, H. (2015). Continuous sign language recognition: Towards large vocabulary statistical recognition systems handling multiple signers. *Computer Vision and Image Understanding*, *141*, 108-125.

[18] Sandler, M., Howard, A., Zhu, M., Zhmoginov, A., & Chen, L. C. (2018). Mobilenetv2: Inverted residuals and linear bottlenecks. In *Proceedings of the IEEE conference on computer vision and pattern recognition* (pp. 4510-4520).

[19] Dong, L., Xu, S., & Xu, B. (2018, April). Speech-transformer: a no-recurrence sequence-to-sequence model for speech recognition. In *2018 IEEE international conference on acoustics, speech and signal processing (ICASSP)* (pp. 5884-5888). IEEE.

[20] Amodei, D., Ananthanarayanan, S., Anubhai, R., Bai, J., Battenberg, E., Case, C., ... & Zhu, Z. (2016, June). Deep speech 2: End-to-end speech recognition in english and mandarin. In *International conference on machine learning* (pp. 173-182). PMLR.

[21] Graves, A., Liwicki, M., Fernández, S., Bertolami, R., Bunke, H., & Schmidhuber, J. (2008). A novel connectionist system for unconstrained handwriting recognition. *IEEE transactions on pattern analysis and machine intelligence*, *31*(5), 855-868.

[22] Assael, Y. M., Shillingford, B., Whiteson, S., & De Freitas, N. (2016). Lipnet: End-to-end sentence-level lipreading. *arXiv preprint arXiv:1611.01599*.

[23] Huang, D. A., Fei-Fei, L., & Niebles, J. C. (2016). Connectionist temporal modeling for weakly supervised action labeling. In *Computer Vision–ECCV 2016: 14th European Conference, Amsterdam, The Netherlands, October 11–14, 2016, Proceedings, Part IV 14* (pp. 137-153). Springer International Publishing.

[24] Cihan Camgoz, N., Hadfield, S., Koller, O., & Bowden, R. (2017). Subunets: End-to-end hand shape and continuous sign language recognition. In *Proceedings of the IEEE international conference on computer vision* (pp. 3056-3065).

[25] Sandler, M., Howard, A., Zhu, M., Zhmoginov, A., & Chen, L. C. (2018). Mobilenetv2: Inverted residuals and linear bottlenecks. In *Proceedings of the IEEE conference on computer vision and pattern recognition* (pp. 4510-4520).

[26] Pan, T. Y., Lo, L. Y., Yeh, C. W., Li, J. W., Liu, H. T., & Hu, M. C. (2018). Sign language recognition in complex background scene based on adaptive skin colour modelling and support vector machine. *International Journal of Big Data Intelligence*, *5*(1-2), 21-30.

[27] Pugeault, N., & Bowden, R. (2011, November). Spelling it out: Real-time ASL fingerspelling recognition. In *2011 IEEE International conference on computer vision workshops (ICCV workshops)* (pp. 1114-1119). IEEE.

[28] Li, D., Yu, X., Xu, C., Petersson, L., & Li, H. (2020). Transferring cross-domain knowledge for video sign language recognition. In *Proceedings of the IEEE/CVF Conference on Computer Vision and Pattern Recognition* (pp. 6205-6214).

[29] Liang, Z. J., Liao, S. B., & Hu, B. Z. (2018). 3D convolutional neural networks for dynamic sign language recognition. *The Computer Journal*, *61*(11), 1724-1736.

[30] Jiang, S., Sun, B., Wang, L., Bai, Y., Li, K., & Fu, Y. (2021). Skeleton aware multi-modal sign language recognition. In *Proceedings of the IEEE/CVF conference on computer vision and pattern recognition* (pp. 3413-3423).

[31] Koller, O., Zargaran, O., Ney, H., & Bowden, R. (2016). Deep sign: Hybrid CNN-HMM for continuous sign language recognition. In *Proceedings of the British Machine Vision Conference 2016*.

[32] Liu, Z., Lin, Y., Cao, Y., Hu, H., Wei, Y., Zhang, Z., ... & Guo, B. (2021). Swin transformer: Hierarchical vision transformer using shifted windows. In *Proceedings of the IEEE/CVF international conference on computer vision* (pp. 10012-10022).

[33] Fan, H., Xiong, B., Mangalam, K., Li, Y., Yan, Z., Malik, J., & Feichtenhofer, C. (2021). Multiscale vision transformers. In *Proceedings of the IEEE/CVF international conference on computer vision* (pp. 6824-6835).

[34] Zhou, H., Zhou, W., Zhou, Y., & Li, H. (2020, April). Spatial-temporal multi-cue network for continuous sign language recognition. In *Proceedings of the AAAI Conference on Artificial Intelligence* (Vol. 34, No. 07, pp. 13009-13016).

[35] Zhang, C. L., Wu, J., & Li, Y. (2022, October). Actionformer: Localizing moments of actions with transformers. In *European Conference on Computer Vision* (pp. 492-510). Cham: Springer Nature Switzerland.

[36] Guo, Q., Qiu, X., Xue, X., & Zhang, Z. (2019). Low-rank and locality constrained self-attention for sequence modeling. *IEEE/ACM Transactions on Audio, Speech, and Language Processing*, *27*(12), 2213-2222.

[37] Paszke, A., Gross, S., Massa, F., Lerer, A., Bradbury, J., Chanan, G., ... & Chintala, S. (2019). Pytorch: An imperative style, high-performance deep learning library. *Advances in neural information processing systems*, *32*.

[38] Glorot, X., & Bengio, Y. (2010, March). Understanding the difficulty of training deep feedforward neural networks. In *Proceedings of the thirteenth international conference on artificial intelligence and statistics* (pp. 249-256). JMLR Workshop and Conference Proceedings.

[39] Deng, J., Dong, W., Socher, R., Li, L. J., Li, K., & Fei-Fei, L. (2009, June). Imagenet: A large-scale hierarchical image database. In *2009 IEEE conference on computer vision and pattern recognition* (pp. 248-255). Ieee.

[40] Kingma, D. P., & Ba, J. (2014). Adam: A method for stochastic optimization. *arXiv preprint arXiv:1412.6980*.

[41] He, K., Zhang, X., Ren, S., & Sun, J. (2016). Deep residual learning for image recognition. In *Proceedings of the IEEE conference on computer vision and pattern recognition* (pp. 770-778).

[42] Min, Y., Hao, A., Chai, X., & Chen, X. (2021). Visual alignment constraint for continuous sign language recognition. In *Proceedings of the IEEE/CVF International Conference on Computer Vision* (pp. 11542-11551).

[43] Zuo, R., & Mak, B. (2022). C2slr: Consistency-enhanced continuous sign language recognition. In *Proceedings of the IEEE/CVF Conference on Computer Vision and Pattern Recognition* (pp. 5131-5140).

[44] Chen, Y., Zuo, R., Wei, F., Wu, Y., Liu, S., & Mak, B. (2022). Two-stream network for sign language recognition and translation. *Advances in Neural Information Processing Systems*, *35*, 17043-17056.